# Touch Analysis: An Empirical Evaluation of Machine Learning Classification Algorithms on Touch Data

Melodee Montgomery[1], Prosenjit Chatterjee[2], John Jenkins[3] and Kaushik Roy[4]

[1,2,3,4] Dept. of Computer Science, North Carolina A&T State University, USA
[1]msmontgomery@aggies.ncat.edu, [2]pchatterjee@aggies.ncat.edu, [3]jmjenki1@aggies.ncat.edu,
[4] kroy@ncat.com

**Abstract.** Our research aims at classifying individuals based on their unique interactions on the touchscreen-based smartphones. In this research, we use 'TouchAnalytics' datasets, which include 41 subjects and 30 different behavioral features. Furthermore, we derived new features from the raw data to improve the overall authentication performance. Previous research has already been done on the TouchAnalytics datasets with the state-of-the-art classifiers, including Support Vector Machine (SVM) and k-nearest neighbor (kNN) and achieved equal error rates (EERs) between 0% to 4%. Here, we propose a novel Deep Neural Net (DNN) architecture to classify the individuals correctly. The proposed DNN architecture has three dense layers and used many-to-many mapping techniques. When we combine the new features with the existing ones, SVM and k-NN achieved the classification accuracies of 94.7% and 94.6%, respectively. This research explored seven other classifiers and out of them, decision tree and our proposed DNN classifiers resulted in the highest accuracies with 100%. The others included: Logistic Regression (LR), Linear Discriminant Analysis (LDA), Gaussian Naive Bayes (NB), Neural Network, and VGGNet with the following accuracy scores of 94.7%, 95.9%, 31.9%, 88.8%, and 96.1%, respectively.

## 1      Introduction

Touch data has become a pivotal part of modern technology. Almost every electronic device that is used today has adopted a touch component. With this growing trend of technology, it is safe to say that the security needs to be updated as well. Biometric authentication is a security process that relies on the unique biological characteristics of an individual to verify that they are indeed the person who they say. It used for identification and surveillance purposes. There are two types of authentication characteristics: physical and behavioral. Physical characteristics include fingerprinting, iris and face scanning, and veins. Most devices use physical characteristics as an authentication method. Fingerprinting, iris scanning, and face recognition are more accurately and easily identified [1]. They virtually take no time to identify considering the scanners are not sabotaged by something such as dirty fingers or smudges preventing it from detecting the user. There are, however, disadvantages of physical characteristics as well. For example, with fingerprints, any skilled hacker with the proper resources would be able to lift a fingerprint from the user and infiltrate the device. For iris and face scanning, any high definition picture of the user could be used as a spoof and trick the scanner. As long as the picture is clear and gets full coverage of the face or the eyes, a hacker could use it as a way to access user information.

Behavioral characteristics include handwriting, voice recognition, type rhythm, and more recently touchalytics. Behavioral authentications are more complicated because behaviors can be difficult to register or easily spoofed. For example, to register voice recognition and to access the device, the user's surroundings must be completely quiet in order not to pick up background noises [1] or falsely register the wrong voices. Also, voices, unlike fingerprints or iris', are liable to change. They can get deeper or, in sickness, raspy or hoarse. The microphone may no longer recognize the user and it would be difficult to access their device. It could also be easy for another user to hack simply by recording the user's voice and replaying it for the device. Handwriting and type rhythm could be duplicated or unreliable. Touchalytics, however, is a fairly new behavioral characteristic that can be used to continuously authenticate the user [2].

This paper serves as an in-depth evaluation on the theoretical successes of using touchalytics as a stand-alone biometric authenticator [2]. Touchalytics uses a person's interaction with their touch screen to discover a unique pattern. Just like a fingerprint, touchalytics is exclusive to a specific person [2] and depending on the features used, it could possibly be one of the safest, most accurate authentication identifiers in the category of biometrics.

## 2    Related Work

The study of touchalytics has been conducted by many researchers. For the past decade, scientists have discovered new ways to incorporate it into the behavioral authentication studies and the effects it has on modern technological security [2, 3, 4]. This technique has yet been formally introduced as a formidable adversary to other, more popular, authenticators such as fingerprinting, iris scanning, or voice recognition.

Touchalytics would be a beneficial addition to the security concerns there are behind the biometric systems in place because they add an extra level of security [5]. Unlike the traditional authenticators, touchalytics is used as a continuous authenticator [1]. This means that, even though a fingerprint could be used to gain initial access to a phone, once that device is unlocked, the phone, along with its data, is accessible to whomever is using it. Touch-based authentication is an inadvertent way of alerting the security system who is currently controlling the device. If the swipe pattern is different than what the phone has registered as the administrative user, it will send a signal without informing the hacker[5, 6] and proceed accordingly by either locking up or limiting usage.

According to research done by Mario Frank et al. [2], touchalytics extracts temporal features of human gestures on planar surfaces to determine individual characteristics of android phones. The purpose of the study was to determine whether a classifier could continuously authenticate users based on the way they interact with the touchscreen of a smartphone. Using over 30 different features and two classifiers, support vector machine (SVM) with a Radial-Basis Function (rbf) kernel and k-nearest-neighbor (*k*-NN), the studies determined that touchalytics alone was not enough to serve as a long-term authenticator. However, it could be used to implement an extended screen-lock time or to augment the current authentication process [2].

In [6], Y. Meng et al. approached their research on touch data by providing 20 participants with the same android phone (HTC Nexus One) to ensure uniformity. The users were instructed to use the phone as they normally would for in order to gather data. The raw data consisted of 120 10-minute sessions of internet browsing, application downloading, etc. After the data was collected, select features were extracted and classified by five different classifiers: Decision Tree (J48), Naïve Bayes, Kstar, Radial Basis Function Network (RBFN), and Back Propaganda Neural Network (BPNN). Using WEKA, the researchers tested each classifier to get the false acceptance rate (FAR) and false rejection rate for each user and the average error rate for each classifier. It was concluded that the RBFN had the best results with an average error rate of 7.71% as opposed to the other classifiers ranging from 11.5% to 24%. They then used an algorithm to combine the RBFN classifier with the Particle Swarm Optimization (PSO) to improve the performance. This hybrid classifier lowered the RBFN's original average error rate from 7.71% to 2.92% for PSO-RBFN.

In [7], using the same dataset in [6], Y. Lee et al. extracted features to measure the differences between users by classifying them. They investigated by using Deep Beliefs Networks (DBN) and Random Forest (RF) classifiers. The researchers were able to classify the developed stroke and session-based feature vectors from the raw data. Using the vectors, they could successfully identify and verify users by applying them to the DBN and RF classifiers. The RF classifier achieved a five-fold cross validation accuracy of 86.5% compared to the DBN accuracy of 81.5% . It was also reported that the DBN was much slower than the RF classifier. They concluded that the DBN was slightly outperformed by the RF but resulted with higher accuracies for certain users causing the results to be inconclusive and required further investigation.

The researchers [8, 9, 10, 11, 12, 13, 14, 15, 18, 19], investigated already in several ways on the reliability and applicability of active authentication by using users' interaction with touchscreens. They measured various types of touch operations, operation lengths, application tasks, and various application scenarios and tried to propose several algorithms that was implemented for classification [18, 19]. Classification used to authenticate the users included nearest neighbor, neural network, support vector machine, and random forest. Results showed equal error rates as low as 1.72% for different types of touch operations with an improvement in accuracy for long observation or shorter training and testing phase timespans.

This paper conducts an empirical evaluation of different machine learning algorithms on touch data. In this research, we find that using several different classifiers will show that although the SVM and k-NN classifiers gave low equal-error rates (EER) [2], other classifiers produce accuracy results that are even better. It is also essential to distinguish the most important features from the touch data; this research applies a Genetic Algorithm (GA) for feature selection. We will use both raw data and the extracted features for classification.

## 3  Methodology

Similar to the deep belief networks (DBN) applied in [7],, this research proposes a deep neural net architecture (DNN) that can efficiently classify the touch analytics on dataset. Our DNN architecture contains build on four dense layers as shown in the Fig.1 and Fig.2. First dense layer will take the input from the input datasets and can handle 10500 parameters, and the output shape has the sub matrix of (None, 3000). First layer has one-to-many mapping with the second dense layer. Second dense layer can handle the parameters of 3001000 and sub matrix of output shape is (None, 1000). Second dense layer has many-to-one and many-to-many mapping with the third dense layer. Third dense layer can process 300300 parameters and has output shape of (None, 34). Forth dense layer has 10,234 parameter capacity. In total, our proposed DNN can handle 3,416,534 parameters and can train them all. Although, we can make the changes and increase or decrease the layer structure according to our dataset volume and features availability. The proposed authentication model is shown in Fig1.

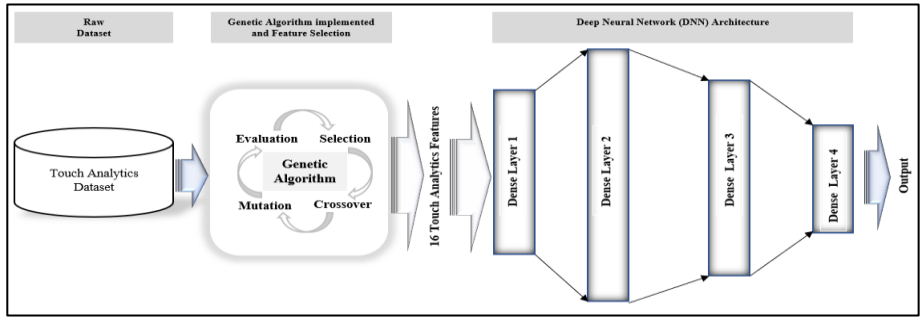

Fig1: Genetic Algorithm and Deep Neural Network architecture for classifying Touch Analytics datasets.

| Layer (Type) | Output Shape | Parameter Handling Capacity |
|---|---|---|
| Dense Layer (dense_1) | (None, 3000) | 1, 05, 000 |
| Dense Layer (dense_2) | (None, 1000) | 30, 01, 000 |
| Dense Layer (dense_3) | (None, 300) | 3, 00, 300 |
| Dense Layer (dense_4) | (None, 34) | 10, 234 |
| Total Parameters: 3, 416, 534 | | |
| Trainable Parameters: 3, 416, 534 | | |
| Non - Trainable Parameters: 0 | | |

Fig2: Dense Layer Structure and their parameter handling capacity for our proposed Deep Neural Network (DNN).

This paper uses a Genetic Algorithm (GA) to select the most important features. The proposed GA uses Charles Darwin's theory of "natural selection" as an optimization technique [16]. It repeatedly modifies a population of individual solutions by operating on a population of artificial chromosomes. The chromosomes have genes and are *fit* with a number to measure the probability of it being a valuable solution to a particular problem. The higher the fitness value is, the more likely the genes chosen will produce the most profitable results [17]. The highest-ranking solutions are then inserted into a *mating pool* where they become *parents*. Two parents, that are chosen at random, produce two *offspring* are expected to generate better quality results than that of the parents.

Next, we conducted an empirical analysis of the different state of the art classification techniques, such as *k*-NN, and SVM with a RBF kernel. To analyze further, with the touch-analytics datasets we experimented on classification through Linear Regression (LR), Decision Tree (CART), Gaussian Naïve Bayes (NB).

## 4    Datasets Used

In this research we used two sets of data. Both datasets featured 41 different users experimenting with 5 android phones (2 of which were the same brand conducted by different experimenters). One dataset contained a combination of over 21,000 instances and over 30 features; this one is referred to as the extracted data. Another dataset used has over 900,000 instances with only 11 features; this is referred to as the raw data. The experiments conducted instructed the users to use the phones, as they naturally would, to read articles and do image comparisons [2]. These experiments served to monitor how the different users interact with everyday usage of touch screen devices. This would help to determine navigational strokes in a usual way so that the study could be used as a continuous security measure. Each individual's touch stroke represented a behavioral attribute similar to that of a fingerprint or signature. No two people operate touch phones the exact same way [4]. After conducting the experiment with SVM and *k*-NN classifiers, the results concluded with an equal error rate of 0%-4%. Although results were fairly good, the researchers determined this conclusion was not enough to use touch screen input as a behavioral biometric alone for long-term authentication [2]. It would however be beneficial to use as an extended security measure for lock-screen timeout.

Studying the subjects required including several different features to maximize the accuracy. The features included in the raw data were the user id, which phone type the user had, the document being examined, time (millisecond) of recorded action, action performed (touch down, touch up, move), phone orientation (landscape or portrait), x-coordinate, y-coordinate, pressure, area covered, and finger orientation.

In addition to testing the raw features, 30 features of the extracted dataset were: inter stroke time, stroke duration, start x, start y, stop x, stop y, direct end to end distance, mean resultant length, up down left right flag, direction of end to end distance, phone ID, 20% 50% and 80% pairwise velocity, 20% 50% and 80% pairwise acceleration, median velocity last 3 points, largest deviation from end to end line, 20% 50% and 80% deviation from end to end line, line average direction, length of trajectory, ratio end to end distance, length of trajectory, average velocity, median acceleration at 1st 5 points, mid-stroke pressure, mid-stroke area covered, mid-stroke finger orientation, change of finger orientation, and phone orientation. These extra features are helpful because the more data gathered, the better the device's chances are at learning the users [5]. Out of the 34 extracted features, we were able to use our proposed GA to predict which of these features was the most important and beneficial to the anticipated results. This is represented in a separate GA dataset for further experimentation.

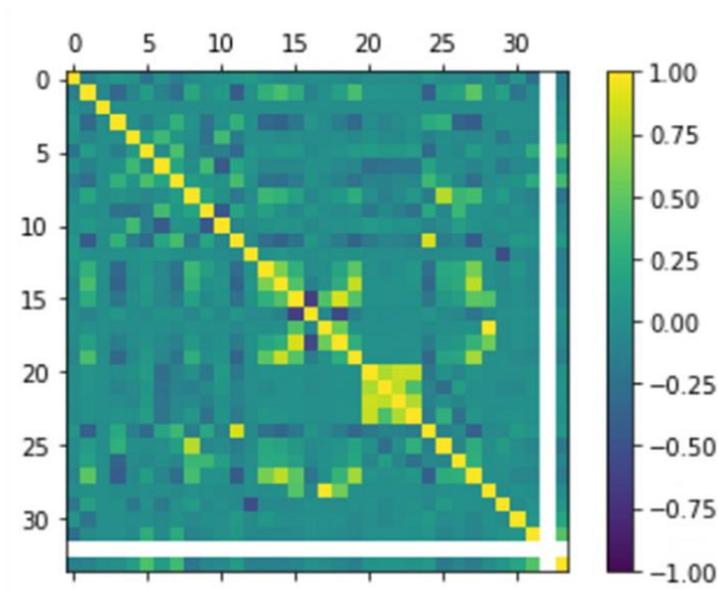

Fig3: Correlation graph of 34 features from extracted dataset

## 5   Results and Discussions

In this paper we proposed an approach to implement touchalytics on datasets for classification purposes. The dataset we received are of two kind; one has 11 features, and the other has 34 features. Here we propose the GA to select the feature subsets from the original dataset and can boast the classification accuracies thereafter. Our proposed

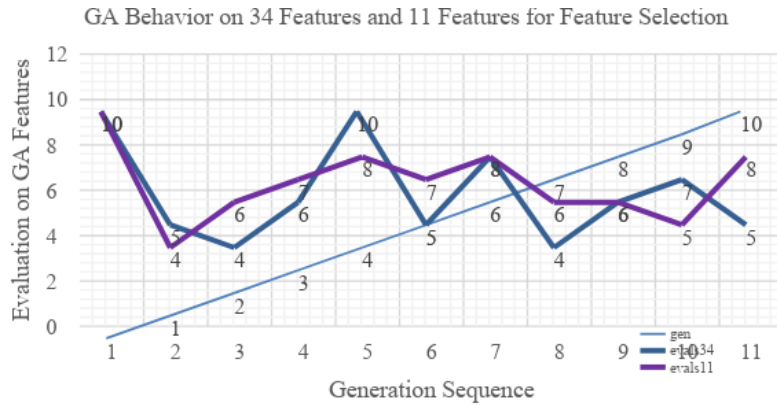

Fig4: Genetic Algorithm implemented on 34 Features and 11 Features Touch Analytics Raw Data.

GA can identify the most selective features that has highest impact on classification accuracies. Figure 4 shows the process of the GA on both the raw and extracted datasets.

We noticed that for the relatively small datasets containing 34 features, when fed into the genetic algorithm, we got a 16-feature subset out of it. Also, from the large dataset with 11 features, we got 7 feature subsets. It is predicted that the Genetic Algorithm will have the better classification results after at least three generation sequences. This is the case because there are many different features to choose from as parents but have low effects on the classification accuracies.

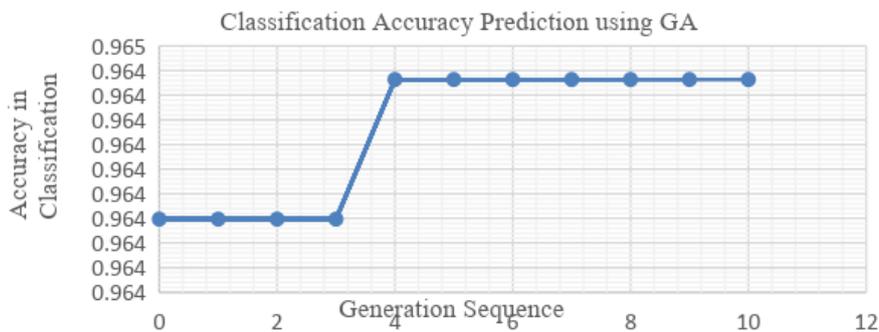

Fig5: Classification accuracy prediction using Genetic Algorithm

After conducting the tests for the datasets, there was about a 90% success rate for the nine classifiers. The only classifier that resulted with a consistently low accuracy was the Gaussian Naïve Bayes classifier. This was expected because the Naïve Bayes' assumption is that the features are independent. According to Figures 4 and 5, many of the features depend on one another and are correlated. The best results we achieved were from the Decision Tree and the "uniquely structured" Deep Neural Net for the extracted dataset. The detailed results from each classifier used on the raw, extracted, and proposed GA datasets are presented in Table 1.

Classification Accuracies on Touchalytics Datasets

| Classifier | Accuracy in Percentage (%) | | | |
|---|---|---|---|---|
| | 11 Raw Features | | 34 Raw Features | |
| | Raw Data | GA Extracted Data (3 Features from 11 Raw Features) | Raw Data | GA Extracted Data (16 Features from 34 Raw Features) |
| Logistic Regression (LR) | 71.00 | 70.70 | 94.70 | 94.10 |
| Linear Discriminative Analysis (LDA) | 77.00 | 77.00 | 95.90 | 94.10 |
| K-neighbors (kNN) | 72.00 | 72.00 | 94.60 | 94.00 |
| Decision Tree (CART) | 79.00 | 79.40 | 100.00 | 100.00 |
| Gaussian Naive Bayes (NB) | 26.00 | 30.00 | 31.90 | 35.00 |
| Support Vector Machine (SVM) | 67.00 | 71.00 | 94.70 | 95.00 |
| VGG Net (VGG) | 89.00 | 91.00 | 96.10 | 96.50 |
| Deep Neural Net (DNN) | 92.00 | 95.00 | 100.00 | 100.00 |

Table1: Comparison of Classification Accuracy on different classifiers along with our proposed Deep Neural Network (DNN) for 11 Features and 34 Features Raw Touch Analytics Data vs. Genetic Algorithm extracted 3 Features (from 11 Features) and 16 Features (from 34 Features).

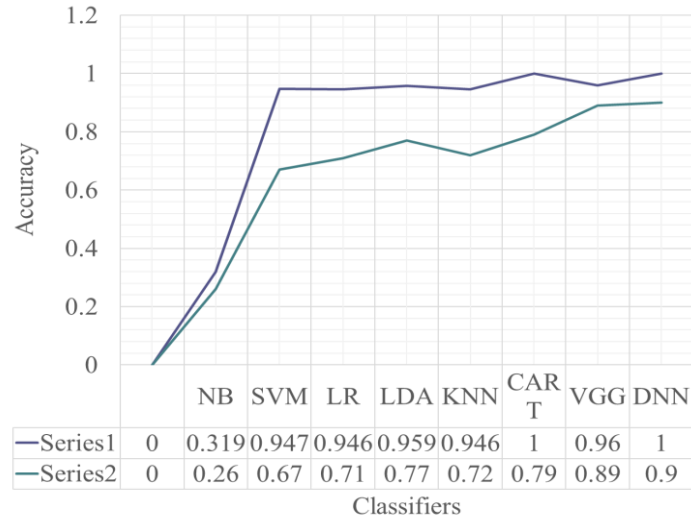

Figure 6: Comparison of classification accuracy based on Extracted Data (series 1) and Raw Data (series 2)

As shown in the table and figure above, our proposed and implemented DNN had the highest accuracies and outperform other classifiers, even the 'modified VGGNet'.

## 6     Conclusions and Future Work

This paper served as an empirical evaluation for the machine learning classification algorithms on the touch analytics datasets presented by Mario Frank and colleagues [2]. We were able to confirm that by running the SVM and $k$-NN classifications on the raw data, the results were favorable. Rather than looking for an Equal Error Rate however, we found the accuracies for these classifiers in addition to others on the raw data and the extracted features data.

By implementing the proposed GA to both datasets, we were able to yield the best possible results for each classifier and compare them to one another. The Decision Tree classifier and our implemented DNN classifier resulted in the highest accuracies with 100% for both the 34 extracted features dataset and also the GA's 16 extracted features dataset. The Gaussian Naïve Bayes Classifier consistently gave low results across the board with 26%, 30%, 31.9%, and 35% for the raw (11 features), GA raw (3 features), extracted (34 features), and GA extracted (16 features) datasets respectively. To further improve touchalytics studies, we plan to conduct experiments using all classifiers on each individual user to get independent classification accuracies.

## 7 Acknowledgements

This research is based upon work supported by the Science & Technology Center: Bio/Computational Evolution in Action Consortium (BEACON) and the Army Research Office (Contract No. W911NF-15-1-0524).